\documentclass[letterpaper, 10 pt, conference]{ieeeconf}  

\IEEEoverridecommandlockouts                              

\usepackage{graphics} 
\usepackage{amsmath} 
\usepackage{amssymb}  

\usepackage[linesnumbered,ruled,vlined]{algorithm2e}
\usepackage{mathtools}
\usepackage{tabularx}
\usepackage{array}
\usepackage{booktabs}
\usepackage{makecell}
\usepackage{adjustbox}
\usepackage{multirow}
\usepackage{url}
\usepackage{flushend}
\title{\LARGE \bf
Efficient Object Detection in Autonomous Driving using Spiking Neural Networks: Performance, Energy Consumption Analysis, and Insights into Open-set Object Discovery}

\author{Aitor Martinez Seras$^{1}$, Javier Del Ser$^{1,2}$ and Pablo Garcia-Bringas$^{3}$
\thanks{Aitor Martinez Seras is a grantee of the BIKAINTEK PhD programme promoted by the Basque Government (ref. 008-B2/2021). Javier Del Ser acknow\-ledges funding support from the same institution through the ELKARTEK programme (EGIA project, KK‑2022/00119).}
\thanks{$^{1}$Aitor Martinez Seras and Javier Del Ser are with TECNALIA, Basque Research \& Technology Alliance (BRTA), 48160 Derio, Spain.         {\tt\small \{aitor.martinez,javier.delser\}@tecnalia.com}}%
\thanks{$^{2}$Javier Del Ser is also with the Dept. of Communications Engineering, University of the Basque Country (UPV/EHU), 48013 Bilbao, Spain}
\thanks{$^{3}$Pablo Garcia-Bringas is with Deusto University, 48007 Bilbao, Spain}
}

\begin{document}

\maketitle
\thispagestyle{empty}
\pagestyle{empty}

\begin{abstract}

Besides performance, efficiency is a key design driver of technologies supporting vehicular perception. Indeed, a well-balanced trade-off between performance and energy consumption is crucial for the sustainability of autonomous vehicles. In this context, the diversity of real-world contexts in which autonomous vehicles can operate motivates the need for empowering perception models with the capability to detect, characterize and identify newly appearing objects by themselves. In this manuscript we elaborate on this threefold conundrum (performance, efficiency and \emph{open-world learning}) for object detection modeling tasks over image data collected from vehicular scenarios. Specifically, we show that well-performing and efficient models can be realized by virtue of Spiking Neural Networks (SNNs), reaching competitive levels of detection performance when compared to their non-spiking counterparts at dramatic energy consumption savings (up to 85\%) and a slightly improved robustness against image noise. Our experiments herein offered also expose qualitatively the complexity of detecting new objects based on the preliminary results of a simple approach to discriminate potential object proposals in the captured image. 
\end{abstract}

\section{Introduction}\label{sec:Intro}

There is wide consensus on the pivotal role of autonomous driving as one of the main challenges of modern Intelligent Transportation Systems (ITS). Among all technologies involved in this paradigm, vehicular perception is a vital functionality to support the awareness and intelligence of the autonomous vehicle. In general, vehicular perception can be defined as the ability of vehicles to sense and understand their surroundings, which is realized by collecting data from various sensors such as LIDAR and automotive cameras \cite{wang2021research}, the latter being the most mainstream sensor for autonomous vehicles. Modeling tasks relying on data collected from these devices are usually approached by using Machine Learning methods, endowing the vehicle with the capability to understand its context and make safe and informed decisions \cite{muhammad2020deep}.

Among such modeling tasks, the related literature has clearly inclined towards semantic segmentation and object detection to model vehicular data. On one hand, semantic segmentation consists of partitioning a scene into various meaningful regions, usually by labeling each pixel in the image, so that each region represents a given semantic concept learned by the model (e.g. pedestrian, vehicle, traffic sign). On the other hand, object detection addresses the problem of identifying and localizing multiple objects in a scene based on a predefined set of object classes of interest for the application at hand \cite{feng2020deep}. 

In this work we focus on the latter (object detection and identification). The upsurge of advances in Machine Learning and mainly Deep Learning (DL) in the last decade has made this family of data-based models dominate the object detection landscape in recent times. Indeed, most methodological contributions for object detection in the last years are based on DL approaches that embed Convolutional Neural Networks (CNNs), pyramid networks or attention mechanisms, to mention a few. Such proposals have elevated object detection to unprecedented levels of performance, as evinced by comprehensive surveys on this topic \cite{zou2023object}.

Unfortunately, despite the reported effectiveness of DL techniques for this modeling task, two major problems still prevail in the object detection research area that hinder the adoption of these methods in real-world vehicular environments. The first issue is the dynamism and unpredictability of vehicular environments. Object detectors typically depart from a set of known classes predefined a priori, thereby considering any unknown object as part of the \emph{background} and thereby, discarded as a potentially identifiable object instance. As a consequence of this predefined search bias, object detectors are not prepared to detect new knowledge autonomously (i.e., new objects that appear recurrently in the vehicular scene) \cite{singh2021order}.

The second challenge emerges from the high power consumption required by these models. Avant-garde DL-based object detectors comprise a huge number of parameters, which require a large amount of computational resources at both training and inference time. This inevitably results in a high energy consumption. Since autonomous vehicles are mobile and self-powered, it is imperative to develop new modeling strategies that deliver enough computing power at a reduced energy consumption \cite{liu2019edge}. This issue becomes more relevant in severely energy-constrained devices (e.g. drones).

In this sense, some recent works have proven the huge efficiency gains and competitive performance of bio-inspired DL models when used for autonomous vehicles. A representative work is the concept of \emph{neural circuit policies} proposed in \cite{lechner2020neural}. A biological source of inspiration also underlie the definition of Spiking Neural Networks (SNNs), which are bio-plausible DL models that provide large energy savings when implemented in specialized neuromorphic hardware. This efficiency comes as a consequence of their event-based nature and the use of time spikes to represent and process information, mimicking the neural activity in the brain. 

Bearing the above in mind, this work aims to explore the potential of this class of biologically inspired models (SNNs) to support efficient and autonomous object detection in vehicular environments. Specifically, we devise a fully-spiking SNN-based object detector based on the well-known non-spiking Faster-RCNN architecture, which is not only modified to learn from temporally encoded data, but also endowed with a simple approach to detect new knowledge at inference time. Experiments are performed over several vehicular perception datasets from the state of the art, yielding qualitative measures that show the competitive performance and large energy efficiency gains achieved by our proposed approach. We complement our experiments with a qualitative assessment of a naive approach to detect new objects based on the information produced by the proposed object detector. Our discussions held in this regard will unveil the difficulty and challenges of detecting new objects with current object detection architectures that are biased towards detecting a set of classes defined beforehand.

The rest of this manuscript is structured as follows: Section \ref{sec:RelatedWork} frames this contribution within the state of the art in object detection, open-world learning and SNNs, whereas Section \ref{sec:proposed} elaborates on the design of the proposed SNN-based detector. Experiments are presented and discussed in Section \ref{sec:ExpSetup}. Section \ref{sec:Conclusions} ends the paper with an outlook towards the challenges emerging from our results and findings.

\section{Related Work}\label{sec:RelatedWork}

This section is divided into four parts. First, Section \ref{ssec:RW_ObjDetection} shortly revisit object detection models proposed by the community to date. Next, Section \ref{ssec:RW_OWOD} describes what open-set or open-world learning refers to, connecting them to the object detection task targeted in this work. Section \ref{ssec:RW_Spiking} pauses at the fundamentals and known problems of SNNs. Finally, Section \ref{ssec:contrib} highlights the contribution of this work.

\subsection{Object detection}\label{ssec:RW_ObjDetection}

As previously mentioned in the introduction, object detection is a computer vision task aimed to detect and locate instances of semantic objects of certain classes in digital images and videos \cite{jiao2019survey}. In the last decade, DL approaches to scene understanding have dominated the state-of-the-art \cite{9913352}, encompassing two main categories: \emph{one-stage} and \emph{two-stage} detectors. In both cases, the architecture is composed by a backbone (usually a CNN) that creates rich feature representations of input images; and a second part that processes those features to perform predictions, which is addressed differently by these two detector types.

To begin with, two-stage detectors follow a coarse-to-fine processing pipeline, dividing the prediction process of the objects in the image into two parts. First, a Region Proposal Network (RPN) realized by a CNN creates a set of potential bounding boxes called \emph{proposals} that potentially contain objects of interest. Then, a detector -- usually composed by several dense layers -- refines the bounding boxes of such proposals output by the RPN, and classifies them into the known categories. This architecture enhances Fast RCNN \cite{fastrcnn2015} by proposing the RPN and creating Faster RCNN \cite{ren2015faster}, a fully DL-based object detector. Later, Feature Pyramid Networks (FPNs) were proposed, a big leap for existing approaches as it devised a top-down architecture with lateral connections that leverages the different spatial resolutions created by CNN backbones in its forward propagation. The idea is to extract feature maps at different depths of the backbone to enable better prediction across a wide variety of scales. The concept of the FPN is a basic building block for many of the detectors contributed ever since.

In contrast, one-stage detectors do not produce any proposals, but directly generate candidate bounding boxes and predictions within a single processing step. Their operating speeds are higher, but their performance degrades notably when detecting dense and small objects \cite{zou2023object}. The first and most used one-stage detector is the so-called \emph{You Only Look Once} (YOLO), which has released several updates over the years to reach its 8th version (Yolov8 \cite{terven2023comprehensive}). Other interesting architectures include the use of transformers as DETR \cite{transformers2020}, an end-to-end transformer-based object detection approach.

\subsection{Open-world assumption}\label{ssec:RW_OWOD}

One of the key assumptions of DL models capable of object detection is the \emph{closed-world} assumption which, when formulated in the context of classification tasks, establishes that all classes to be detected appear in the training set. The emergence of new classes can compromise the performance of models in the open world, motivating research for different tasks addressing this problem. 

Object detection is not an exception to this statement: specifically, \emph{open-set} object detection seeks to solve the problems of overconfident classification of unknown objects as one of the known classes. Several previous studies have tackled this problem: among them, in \cite{miller2018dropout} Monte Carlo sampling is used to estimate the uncertainty associated to predictions, such that a high uncertainty is declared to be an indication of unknown inputs to the model. More recently, \cite{dhamija2020overlooked} was the first to formalize a protocol to measure the open-set capabilities of models. 

Further along this line, some other works define and elaborate on what is called \emph{Open-World Object Detection} (OWOD), whose goal is not only to classify unknown objects as such, but also to characterize and incorporate them into the knowledge of the detector incrementally (i.e. without retraining). Works related to OWOD include \cite{openworldobjectdetection2021}, which designed a strategy based on two-stage detectors, an unknown auto-labeling method and a contrastive clustering phase to detect new objects. In essence, they maintain a prototype vector for each class (including the unknown class), and enforce class separation in the feature space by adding a clustering loss that pushes dissimilar classes further apart from each other. Since the unknown class needs to be taught to the model as one of its possible outcomes, and no labels are present in the dataset, they annotate potential unknown objects with the RPN using a simple heuristic: proposals with high objectness that do not overlap with ground truth classes are declared to be unknown objects. To incrementally learn new classes, their strategy is to store a balanced set of exemplars and fine tune them after each incremental step. A similar method is used in \cite{singh2021order}, where the target is to detect unknowns in road scenes. To solve the specific problems of this environment, Feature-Mix is used to enhance the discrimination of unknowns, together with focal regression to tackle the intra-class scale variation of vehicle scenes  and curriculum learning to improve the detection of small objects.

A natural direction departing from OWOD is to design methods to automatically perform the labeling operation. This is precisely what Open-Set Detection and Discovery or Novel Class Discovery and Localization does. The proposed strategy in \cite{towardsopensetobjdetanddiscovery2022} follows up the one proposed in \cite{openworldobjectdetection2021} to discover the unknowns, carrying out K-means and unsupervised mix-up contrastive learning to assign them a new category. Another work addressed the task similarly utilizing RPN to assign pseudo-labels to proposals via online constrained clustering to perform a self-supervised training of an added secondary classification head for the unknowns, where the number of novel classes is defined empirically. 

In summary, there is a clear concern in the literature with the limitations of object detectors when deployed in real-world highly dynamic scenarios, including vehicular scenes.

\subsection{Spiking neural networks}\label{ssec:RW_Spiking}

Often termed as the \textit{third generation of Artificial Neural Networks} (ANNs), SNNs are bio-inspired DL models that process information in the form of temporal binary events or \emph{spikes} across multiple time steps. This asynchronous nature, together with the sparsity of spikes and the fact that energy is only consumed by SNN neurons when spikes are produced, allow for the deployment of these models on highly energy-efficient processing hardware suited to deal with temporally encoded data (\emph{neuromorphic hardware}) \cite{davies2018loihi}\cite{akopyan2015truenorth}. As SNNs operate with spikes, static images must be converted into spike trains by means of an encoding method. Existing methods can be categorized into two types; i) temporal encoding methods, which carry information in the exact timing the spike occurs; and ii) rate-based methods, which convey the information in the rate at which spikes are generated. In the latter, several mainstream methods exist, including Poisson rate encoding \cite{heeger2000poisson} or direct input encoding \cite{rathi2021diet}.

For the activation function, several neuron models have been developed, ranging from very biologically plausible but computationally heavy ones like the Izhikevich model \cite{izhikevich2003} to more simple and computationally cheap as the Leaky Integrate-and-Fire (LIF) model \cite{kasabov2019time}. These models explain how the neuron's membrane potential evolves and how spikes are generated at varying levels of detail. Essentially, the LIF neuron operates by integrating input spikes as currents into an resistor–capacitor circuit, causing the voltage to increase until it surpasses a threshold. At this moment during the simulation period, a spike is emitted towards the downstream neurons (neurons of the next layer).

SNNs have a significant disadvantage with respect to non-spiking approaches: spiking neurons are non-differentiable due to the spike's discontinuity. As a consequence, gradient backpropagation cannot be directly applied. Several learning techniques have been developed over the years to overcome this problem, including novel bio-inspired learning rules like the Spike-Time Dependent Plasticity (STDP) \cite{caporale2008spike}, Surrogate Gradient methods (SG) and ANN-to-SNN conversion techniques. This last group of techniques first train a traditional ANN, and then translate its weights to a SNN. Unfortunately, they struggle to efficiently capture spike information through the temporal domain, consequently requiring hundreds or thousands of timesteps to maintain the performance level, and ultimately ruining any efficiency gains. In contrast, SG methods directly train SNNs by replacing the discontinuity of the neuron activation function with a smooth continuous surrogate, harnessing the benefits of SNNs. SuperSpike \cite{zenke2018superspike} is one of these methods, which minimizes the van Rossum distance between the desired and the actual spike train via gradient descent.

\subsection{Contribution} \label{ssec:contrib} 

To the best of our knowledge, no prior work has pulsed whether fully-spiking SNN-based object detector can perform competitively when used to detect objects from vehicular image data, neither in terms of detection performance statistics nor in what refers to their energy efficiency. Some works implementing SNNs for scene understanding have been contributed in the recent past, but most of them apply ANN-to-SNN conversion methods \cite{cordone2022object}\cite{kim2020spikingyolo}, use less challenging datasets \cite{kim2022beyond} or tackle simpler tasks like single object location \cite{barchid2022spiking}. Along with the potential of directly training an energy-efficient SNN object detector, this work uncovers the inherent difficulty of detecting new objects due to the bias induced by the predefined set on new classes from which the object detector is learned. A relatively simple approach that does not require any retraining of the model will be designed to buttress our claims in this matter.

\section{Proposed SNN-based detector} \label{sec:proposed}

Once the contribution of the paper has been stated, this section proceeds by detailing the design of the proposed fully-spiking SNN-based detector trained by using SG methods (Section \ref{ssec:ObjDetection}), followed by the description of the simple strategy proposed to endow the object detector with the ability to discriminate new unknown objects (Section \ref{ssec:OWOD}).

\subsection{Direct training of SNNs for object detection}\label{ssec:ObjDetection}

As argued throughout preceding sections, SNNs can unleash huge energy savings thanks to the event-based nature of spikes when implemented in neuromorphic hardware. There are plenty of methods designed to train this kind of networks. Among them, ANN-to-SNN conversion can be considered the most adopted strategy for large model architectures due to its straightforward implementation. The major drawback of this conversion strategy is its lower energy efficiency and its proven difficulties to learn complex temporal dynamics, requiring high computational overheads and a long inference latency to better approximate the ANN weight values \cite{sengupta2019going}. To overcome these downsides, in this work we propose to train a two-stage SNN-based object detector by applying SG methods, which allow maintaining all the benefits of SNNs while barely losing performance w.r.t. traditional ANN detector. Specifically, we show that a fully-spiking Faster RCNN architecture model with FPN can be directly trained over the temporal domain.

\begin{figure*}[ht]
	\centering
	\includegraphics[width=2\columnwidth]{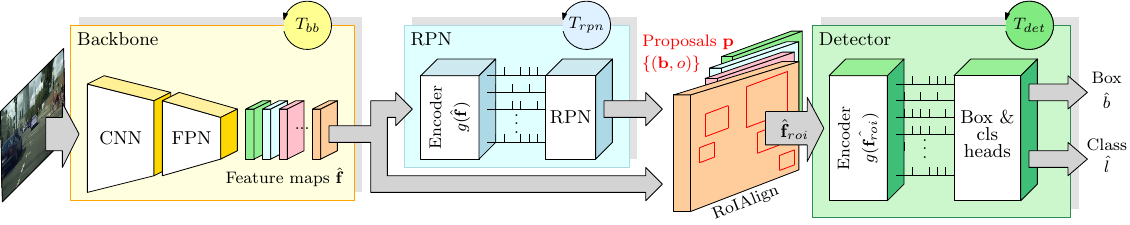}
	\caption{Block diagram of the proposed fully-spiking object detector, which can be directly trained over the time domain. The architecture is divided into 3 blocks (backbone, RPN and detector). Two encoders are used to transform the inputs to the RPN and detector into spike trains.}
	\label{fig:training}
	\vspace{-3mm}
\end{figure*}

The overall design of this object detection model is depicted in Fig. \ref{fig:training}. First it is important to recall that SNNs process data in the temporal domain, hence static data must be converted into time-dependent inputs, so that the processing stage can be done during a certain amount of time (i.e. simulation time). This time is discretized into time steps $\Delta T$, so that neurons can only emit one spike per time step $\Delta T$. Consequently, simulation time is often defined as an integer multiplier of the time step, e.g., $T \in \mathbb{N}$. 

Following Fig. \ref{fig:training}, the overall Faster RCNN architecture is divided into its three main components: 1) the backbone (CNN and FPN), 2) the RPN; and 3) the detector itself. Each part is responsible of providing to the downstream modules with refined information. The backbone extracts rich features out of the images at different spatial depths, whereas the RPN processes the feature maps and generates therefrom proposals that point out regions of the input image where an object is possibly located. Finally, the detector leverages both features and proposals to output refined bounding boxes, together with the predicted class for each box. This three architectural division is key to our proposed method, as the processing inside each part can be done independently of each other, insofar as each module provides the subsequent with the required information. Hence, three different simulation loops can be arranged (one per component), each with its own simulation time (namely, $T_{bb}$, $T_{rpn}$ and $T_{det}$), so that the inference time is given by their sum. They must be executed sequentially and can be trained separately using any preferred SG method, together with Backpropagation through time (BPTT). The loss functions are those of Faster RCNN, i.e.:
\begin{eqnarray}
\mathcal{L}_{rpn} = \mathcal{L}_{obj} + \lambda_{rpn}\cdot \mathcal{L}_{reg} \mbox{ [RPN]},\\
\mathcal{L}_{det} = \mathcal{L}_{cls} + \lambda_{det}\cdot \mathcal{L}_{box} \mbox{ [Detector]},
\end{eqnarray}
where $\mathcal{L}_{obj}$ is a classification loss over 2 classes (\textit{object} or \textit{not object}); $L_{reg}$ is a regression loss of the bounding boxes that is only computed when there is an object; $\mathcal{L}_{cls}$ denotes a log-loss associated to the true object class; and $\mathcal{L}_{box}$ is a bounding box loss computed for non-background objects.

Splitting the architecture in three processing stages also allows for leveraging existing pretrained feature extraction backbones. Currently scarce open source implementations of spiking CNN backbones can be found in the literature: the few existing approaches lack good performance and none of them incorporates a FPN. Hence, we utilize a traditional ANN for the backbone to extract features of the images, without compromising the fully-spiking training process of the overall detector. Nevertheless, fully-spiking feature extractors could be plugged to the system and apply the same training protocol as the one presented in this work. 

The detailed workflow is as follows. First, the backbone extracts features $\hat{\mathbf{f}} \in \mathbb{R}^D$ corresponding to the input image at different resolutions (different \emph{spatial depths} or \emph{levels}). Next. the RPN block receives those feature maps and generates proposals $\mathbf{p} = \{(\mathbf{b}, o)\}$, where $\mathbf{b}\in\mathbb{R}^4$ are the predicted coordinates of the proposal and $o\in\mathbb{R}$ is its \emph{objectness} score. Then a RoI-Align layer \cite{maskrcnn} crops the feature maps $\hat{\mathbf{f}}$ using the proposals $\mathbf{p}$, obtaining fixed sized ROI-aligned features $\hat{\mathbf{f}}_{roi} \in \mathbb{R}^{c\times h \times w}$, where $c$ is the number of input channels, and $h$ and $w$ are hyperparameters. Such aligned features are finally input to the detector, where final predictions $\hat{y} = \{\hat{b}, \hat{l}\}$ are obtained. Predictions include $\hat{b}\in\mathbb{R}^4$, which are the final bounding box coordinates; and $\hat{l}\in\mathbb{R}^{K+1}$, which stands for the softmax class probabilities predicted for the bounding box ($K$ classes + \emph{background}).

Both RPN and detector blocks operate over spikes $s\in\{0,1\}$, i.e. temporal binary values. Therefore, an encoder layer $g(\hat{\mathbf{f}}):~\mathbb{R}^D \mapsto \{0,1\}^{T \times D}$ (see Fig. \ref{fig:training}) must be attached to each of such blocks, converting the incoming features $\hat{\mathbf{f}}$ into spike trains $\mathbf{s}(t)=\{s_t(t)\}_{t=1}^T$, namely, series of spikes that encode the input information in the time domain. A direct input encoding strategy is used to connect each pixel of the feature maps to one LIF neuron with fixed parameters (same parameters for all pixels). Consequently, the value of the pixel can be interpreted as the current intensity arriving to that neuron's membrane at every time step. Real-valued floating-point numbers corresponding to predictions $\mathbf{p}$ and $\hat{y}$ are achieved by placing LI (\emph{Leaky-Integrator}) neurons in the last layer of each block, which only accumulate voltage throughout the entire simulation time $T$. We interpret the last value of the membrane voltage as the prediction, either a logit for a softmax or a bounding box delta for proposal and bounding box regression.

It is interesting to remark that this strategy allows controlling the performance and energy efficiency trade-off driven by the choice of the simulation time $T$. As the number of time steps increases, SNNs are capable of better information processing and hence of potentially improving performance, while also increasing its energy consumption as the result of an increased number of operations. The three different loops devised for the compounding blocks of the architecture, each with its simulation time $T_{bb}$, $T_{rpn}$ and $T_{det}$, grants flexibility to explore the performance-efficiency trade-off for a fixed value of the parameters of the encoding neurons. 

\subsection{Novel object discovery}\label{ssec:OWOD}

One concerning problem of existing object detectors is their incapability to recognize new objects. In this sense, some previous works attempt to achieve this by applying several strategies that involve retraining the model in a supervised manner, adding a new class for unknown objects. The labels for these unknown objects are extracted from the RPN using different methods. In this work, we devise a new technique that does not entail any retraining of the model. However, we do not characterize nor incrementally incorporate unseen classes to the model's knowledge; therefore, we aim to solve a new object discovery problem. 

To this end, we exploit the information that is inherently produced by two-stage detectors like the one in which the proposed model relies (Faster RCNN). At inference time, the detector refines the proposals $\mathbf{p}$ inferred by the RPN, and predicts the object as one of the $K$ classes or as belonging to the \emph{background}. The key idea of our approach is to further exploit \emph{background} detections together with the vast amount of proposals generated by the RPN, to output a \emph{new object} score $\mathbf{Q}$ for each \emph{background} prediction, such that it can be used to discover new unknown objects.

Before diving into the method, we must clarify a few internal steps within the RPN block. After assigning one proposal per anchor box and per pixel of the feature maps, the RPN, independently per spatial depth, retains only $\mathbf{p}_{pre}$ proposals, i.e., the top-$k_1$ proposals by objectness, where $k_1$ is a hyperparameter. Then, it performs Non-Maximum Supression (NMS) separately by spatial level, ranks again the proposals by their objectness but regardless of the level, and finally selects the top-$k_2$ final proposals $\mathbf{p}$, where $k_2$ is another hyperparameter. For out technique, we keep all the $\mathbf{p}_{pre}$ proposals prior to the NMS operation.

The summary of the method devised to discover new objects is presented in Algorithm~\ref{alg:new_obj}. For each image, the $\mathbf{p}_{pre}$ proposals are collected together with their objectness (Line \textbf{1}). From the detector we discriminate all \emph{background} ($bg$) predictions that do not overlap with the predictions of known objects for the image, i.e.: 
\begin{equation}
\hat{\mathbf{y}}^{*}_{bg} = \left\lbrace\hat{y}^{i}_{bg} \forall i\in\{1,\ldots,k_2\} : IoU(\hat{y}^{i}_{bg}, \hat{\mathbf{y}}) = 0\right\rbrace,
\end{equation}
where $IoU(y,\mathbf{y}')$ denotes the intersection over union of the bounding box linked to $y$ and all those bounding boxes of predictions in $\mathbf{y}'$, $\hat{y}^{i}_{bg} = \{(\hat{b}, 0)\}$, and superscript $i$ refers to each of the predictions. We treat these $\hat{\mathbf{y}}^{*}_{bg}\doteq \{\hat{y}^{*, i}_{bg}\}$ as potentially new objects. For each of these non-overlapping predictions, a score $Q^i$ is computed that informs about the possibility that the bounding box of $\hat{y}^{*, i}_{bg}$ has a new object. 
\begin{algorithm}
	\DontPrintSemicolon
	Collect pre-NMS proposals $\mathbf{p}_{pre} = \{(\mathbf{b}_{pre}, o_{pre})\}$ from the RPN block\;
	Extract $\hat{\mathbf{y}}^{*}_{bg} = \{\hat{y}^{i}_{bg}\}^{k_2}_{i=1} : IoU(\hat{y}^{i}_{bg}, \hat{\mathbf{y}}) = 0$ \;
	\For{$\hat{y}^{*, i}_{bg}\in\hat{\mathbf{y}}^{*}_{bg}$}{
		$\mathbf{U}^{i}~=~\{ IoU(\hat{y}^{*, i}_{bg}, \mathbf{b}_{pre}^j)\}^{k_1}_{j=1}$ \;
		$\mathbf{V}^{i}=\{ o^{j}\}^{k_1}_{j=1}$ \;
		$\mathbf{M}^{i} = \mathbf{U}^{i} \odot \mathbf{V}^{i} $ \;
		${Q}^{i} = \sum^{k_1}_{j=1}{M}^{i}_{j}/\left|\{i\}_{i=1}^{k_1}:\: \mathbf{U}^{i} > 0\right|$
	}
	Proposals with new objects: $\left\lbrace\{\hat{y}^{*, i}_{bg}\}_{i=1}^{k_1}:\: Q^i>\gamma_Q\right\rbrace$\;
	\caption{New object discovery}
	\label{alg:new_obj}

\end{algorithm}

To this end, we compute the $IoU$ score of each \emph{background} prediction $\hat{y}^{i}_{bg}$ with all the boxes within the set of collected proposals $\mathbf{p}_{pre}$, hence obtaining $\mathbf{U}^{i}=\{ IoU(\hat{y}^{*, i}_{bg}, \mathbf{b}_{pre}^j)\}^{k_1}_{j=0}$ (Line \textbf{4}). We also collect their objectness loss value, such that $\mathbf{V}^{i}=\{o^{j}\}^{k_1}_{j=0}$ (line \textbf{5}). Then, we weight each of the objectness values with their corresponding $IoU$ by computing the element-wise product, resulting in $\mathbf{M}^{i} = \mathbf{U}^{i} \odot \mathbf{V}^{i}$ (Line \textbf{6}). Finally, we compute a weighted average of these objectness values (Line \textbf{7}), but taking into account only the objectness of the $\mathbf{p}_{pre}$ proposals that actually overlap with the evaluated BG prediction, namely:
\begin{equation}
{Q}^{i} = \frac{\sum^{k_1}_{j=1}{M}^{i}_{j}}{\left|\{i\}_{i=1}^{k_1}:\: \mathbf{U}^{i} > 0\right|},
\end{equation}
where $|\cdot|$ denotes set cardinality. By imposing a threshold $\gamma_{Q}$ on this score, a set of proposals that potentially include new objects is produced (Line \textbf{8}). The intuition behind this strategy is that \emph{background} predictions that do not overlap with known object predictions could contain new objects and that these \emph{background} predictions, if they overlap with proposals characterized by a high objectness, may indicate that an object is contained on them. In addition, it is important to divide the score obtained for each \emph{background} prediction by the number of overlapping proposals to reward having overlapping proposals with great objectness rather than having a large number of overlapping proposals.

\section{Experiments and Results}\label{sec:ExpSetup}

A experimental study has been carried out to answer the following research questions:
\begin{itemize}
	\item \textit{RQ1}: Does a directly trained SNN-based object detector achieve competitive results for vehicular perception? 
\item  \textit{RQ2}: How energy-efficient is the proposed detector?
\item \textit{RQ3}: Is the proposed detector robust against noise?
\item \textit{RQ4}: Can the proposed model detect unknown objects?
\end{itemize}

The datasets used for the evaluation of the performance of the object detection model are \texttt{Cityscapes} \cite{cordts2016cityscapes}, Indian Driving Dataset (\texttt{IDD}) \cite{varma2019idd} and Berkeley Deep Drive (\texttt{BDD}) \cite{yu2020bdd100k}. \texttt{Cityscapes} for object detection contains 3,475 images (2,975 train, 500 val) of several cities of Germany and different 8 classes; \texttt{IDD} includes 15 classes in 31,569 train and 10,225 validation images of Hyderabad and Bangalore; and \texttt{BDD} comprises 70,000 and 10,000 train and validation images of various environmental conditions with 10 classes. 

In our experiments, $2^{nd}$ order LIF neurons are utilized, which consider synaptic conductance, time-decaying voltage and synaptic currents, and the Heaviside step function for the jump condition. We have used the framework Norse \cite{pehle2021norse} for the implementation. Our model is based on the standard Faster R-CNN \cite{fastrcnn2015} with a ResNet-50 plus FPN backbone, pretrained over the CoCo dataset. As the number of train images greatly differs between datasets, we train in each case with different hyperparameters, but sharing the same training protocol. First we trained the RPN freezing the layers of the detector and then we train the detector using the already trained RPN while freezing its layers. Finally, as finetuning FPN brings greater performance, we undertake a final training of FPN, RPN and detector layers with a very low learning rate. We trained non-spiking models with the same architecture and same protocol but different hyperparameters as well, reporting the performance of both fine-tuned and non-fine-tuned models.

\begin{figure}[!h]
    \centering
    \includegraphics[width=0.48\columnwidth]{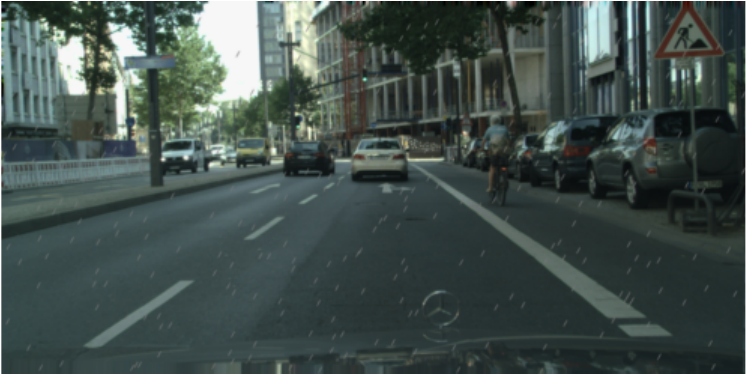}
    \includegraphics[width=0.48\columnwidth]{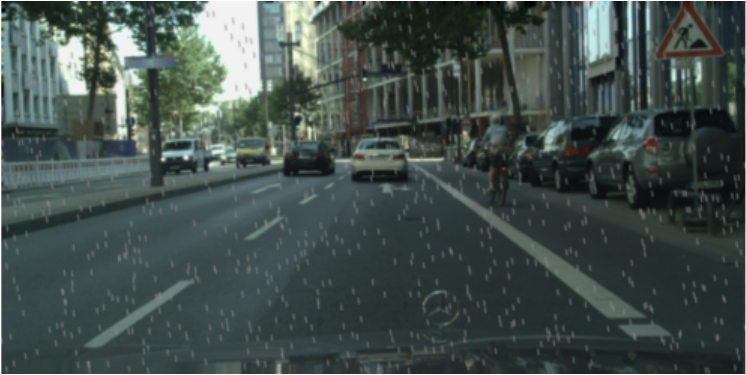}
    \caption{Light (left) and heavy (right) rain simulation.}
    \label{fig:rain}
     \vspace{-3mm}
\end{figure}

We evaluate the performance of known classes in the validation set using the mean Average Precision at an IoU threshold of 0.5 ($\text{mAP}_{.5}$), the $\text{mAP}_{.5:.95}$ as in \cite{lin2014coco} and the mean Average Recall at an IoU threshold of 0.5 ($\text{mAR}_{.5}$). To evaluate the robustness against noisy environments (\textit{RQ3}), two type of noises are artificially induced to the validation images: Gaussian and rain simulating noise. The latter is done by adding both blur and sharp lines mimicking drops and by reducing the brightness of the scene (Fig. \ref{fig:rain}). Source code has been released in \url{https://github.com/aitor-martinez-seras/SNN-Automotive-Object-Detection.git}.

\subsection{RQ1: Does a directly trained SNN-based object detector achieve competitive results for vehicular perception?}\label{ssec:RQ1}

\begin{table*}[!htbp]
\caption{Performance and efficiency results (\%). $\Delta_E\doteq E_{\text{SNN}}/{E_{\text{ANN}}}$ (ratio of the estimated energy consumption of SNN and ANN)} \label{tb:mAP_and_eff}
\centering
\renewcommand\arraystretch{1} 
\resizebox{\textwidth}{!}{\begin{tabular}{ccccccccccccccc}
    \toprule
     & \multicolumn{4}{c}{\texttt{Cityscapes}} & & \multicolumn{4}{c}{\texttt{IDD}} & & \multicolumn{4}{c}{\texttt{BDD}} \\
    \cmidrule{2-5} \cmidrule{7-10} \cmidrule{12-15}
    Model  & $\text{mAP}_{.5}$ & $\text{mAP}_{.5:.95}$ & $\text{mAR}_{.5}$ & $\Delta_E$ & & $\text{mAP}_{.5}$ & $\text{mAP}_{.5:.95}$ & $\text{mAR}_{.5}$ & $\Delta_E$ & & $\text{mAP}_{.5}$  & $\text{mAP}_{.5:.95}$ & $\text{mAR}_{.5}$ & $\Delta_E$ \\
    \midrule
    $\textit{SNN}$ & 50.4 & 27.8 & 37.5 & 48.5  & &  34.3 & 19.5 & 26.3 & 40.1  & & 44.7 & 22.5 & 32.8 & 36.1 \\
    $\textit{SNN}_{\ast}$ (fine-tuned) & 52.3 & 28.3 & 37.8 & 49.1  & &  34.7 & 20.1 & 26.6 & 39.1  & &  45.4 & 23.2 & 33.5 & 36.8 \\
    \midrule
    $\textit{ANN}$ & 50.2 & 27.9 & 37.5 & 100 & &  34.1 & 19.6 & 25.7 & 100 & & 46.4 & 23.6 & 33.8 & 100 \\
    $\textit{ANN}_{\ast}$ (fine-tuned) & 52.1 & 29.4 & 38.1 & 100 & & 35.3 & 20.6 & 26.6 & 100 & &  46.8 & 24.2 & 34.0 & 100\\
    \bottomrule
    \end{tabular}}
    \vspace{-3mm}
 \end{table*}

We compare the performance of the proposed fully-spiking detector (SNN) and its non-spiking version (ANN), obtaining the results in Table \ref{tb:mAP_and_eff}. As expected, the best performance is achieved by non-spiking fine-tuned models (ANN$_{\ast}$), being the gaps to their spiking counterparts (SNN, SNN$_{\ast}$) less than $1.8\%$ in all cases. It is interesting to note that when no fine tuning is done, the performance of the SNN-based detector is slightly higher in the cases of \texttt{Cityscapes} and \texttt{IDD}. In summary, we conclude that the SNN-based detector performs competitively compared to its non-spiking counterparts.
\subsection{RQ2: How energy-efficient is the proposed detector?}\label{ssec:RQ2}

To measure the efficiency we embrace the methodology proposed in previous works \cite{rathi2021diet,kim2022beyond} to calculate the energy consumption of both models under comparison. We first compute the number of floating point operations (FLOPs) per layer for a traditional ANN as:
\begin{equation}\label{eq:flops_ann}
	\text{FLOPs}_{\text{ANN}}(L)=
	\begin{cases}
		O^2 \cdot C_{in} \cdot  k^2 \cdot  C_{out} & \text{if } L = \texttt{conv}\\
		C_{in} \cdot C_{out}  & \text{if } L = \texttt{linear} 
	\end{cases}
\end{equation}
where $O$ is the size of the output feature maps size, $k$ is the kernel size and $C_{in}$ and $C_{out}$ are the number of input and output channels. FLOPs in the SNN yield similarly as:
\begin{equation}\label{eq:flops_snn}
    \text{FLOPs}_{\text{SNN}}(L) = \text{FLOPs}_{\text{ANN}}(L) \cdot T \cdot S_A,
\end{equation}  
where the spiking activity factor $S_A$ -- calculated as the average spike rate per layer over the set of instances under consideration -- accounts for the fact that computations are only done whenever an spike occurs. Finally, the total inference energy across all layers is obtained as:
\begin{align}
	E_{\text{ANN}} & = \sum_L \text{FLOPs}_{ANN}(L) \cdot E_{\text{MAC}}, \label{eq:energy_ann} \\
	E_{\text{SNN}} & = \sum_L \text{FLOPs}_{SNN}(L) \cdot E_{\text{AC}}, \label{eq:energy_snn}
\end{align} 
where the energy per operation is set to $E_{\text{MAC}}=4.6$ pJ and $E_{\text{AC}}=0.9$ pJ for a 45nm processor \cite{computing2014energy}. $E_{\text{MAC}}$ refers to the energy consumed by a multiply and accumulate operation, the type of operation in performed the forward pass of a traditional ANN, while $E_{\text{AC}}$ is the energy consumed in the accumulate operation, what SNNs use.

\begin{figure}[!h]
    \centering
    \vspace{-2mm}
    \includegraphics[width=0.5\columnwidth]{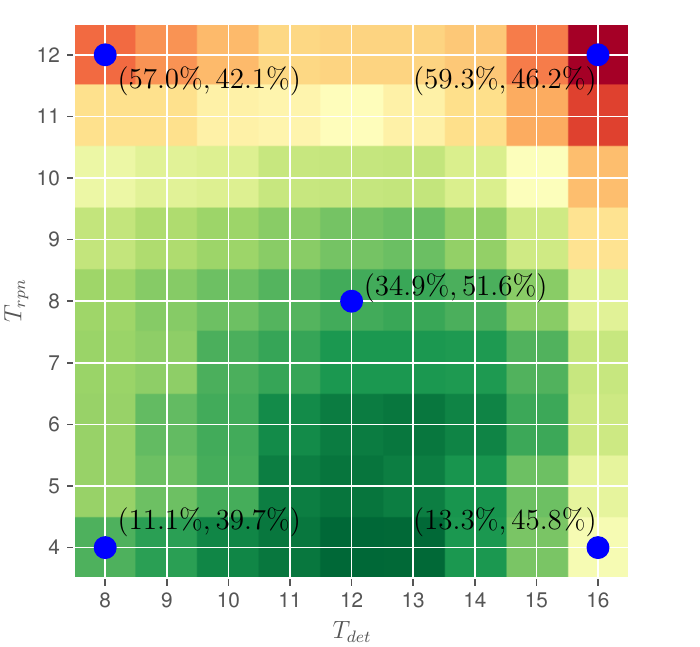}
	\vspace{-2mm}
    \caption{Performance-accuracy trade-off for the proposed detector trained with $T_{rpn} = 8$ and $T_{det} = 12$ when varying $T_{rpn}$ and $T_{det}$ without retraining. The pair of values depicted are ($\Delta_E$[\%], $\text{mAP}_{.5}$ [\%]).}
	\vspace{-3mm}
    \label{fig:tradeof_eff}
\end{figure}

Our methodology provides flexibility to define different number of time steps for each part of the model. Therefore, besides the overall efficiency of the models (Table \ref{tb:mAP_and_eff}) we have also inspected the performance-efficiency trade-off that can be obtained by i) training the model over less time steps (Table \ref{tb:energy_eff}); and 2) by changing the simulation time of an already trained model (Fig. \ref{fig:tradeof_eff}). To begin with, Table \ref{tb:energy_eff}, reveals that the efficiency of the model can be improved by decreasing the simulation time of each module, while maintaining the performance. It is also remarkable  that increasing the number of time steps lead to a degradation in performance together with reduced energy savings. This counter-intuitive behavior may be due to several reasons. First, SG methods use BPTT, so more timesteps can result in vanishing gradients. Another explanation is that SNN training must be done with low batch sizes due to the hardware constraints, which yields unstable gradients and hence a higher variability of the model's convergence. In fact, this effect becomes more acute as we increase $T$, since the hardware bottleneck escalates and increasingly lower batch sizes are needed.
\begin{table}[!htbp]
\caption{Energy consumption reduction vs $mAP_{.5}$ (\texttt{Cityscapes})} \label{tb:energy_eff}
\centering
\renewcommand\arraystretch{1} 
\resizebox{0.5\columnwidth}{!}{\begin{tabular}{cccc}
    \toprule
	\makecell{$T_{rpn}$} & \makecell{$T_{det}$} & $\Delta_{E}$ $\downarrow$ & $\text{mAP}_{.5}$ $\uparrow$\\
    \midrule
    4 & 8 & 15.8\% & 50.7\% \\
    6 & 8 & 25.6\% & 50.4\% \\
    8 & 12 & 34.9\% & 51.6\% \\
    10 & 12 & 41.8\% & 51.0\% \\
    12 & 16 & 48.5\% & 50.4\% \\
    \bottomrule
    \end{tabular}}
\vspace{-3mm}
 \end{table}

The second experiment related to RQ2 aims to elucidate the performance-efficiency trade-off when changing $T$ for different parts of an already trained model. To picture that, we have used the plot shown in Fig. \ref{fig:tradeof_eff}, where green represents models with great accuracy and efficiency while red represents the opposite. It is interesting to note that the RPN has the greatest influence on this trade-off, as decreasing $T_{rpn}$ leads to significant energy savings without compromising the precision of the model. This is clearly depicted by the leftmost blue points, where the lowest $T_{rpn}$ results in $45\%$ energy consumption reduction while only loosing $3\%$ of predictive performance. In contrast, varying $T_{det}$ does not entail great energy savings, while impacting severely on the average precision.

\subsection{RQ3: Is the proposed detector robust against noise?}\label{ssec:RQ3}

Another property often attributed to SNNs is their robustness against noisy inputs, mainly by virtue of temporal encoding processes that transform input values into multiple spikes that follow a certain random distribution, introducing noise to the training process \cite{kim2022beyond}. We have compared the relative precision decrease of the developed SNN-based detector and non-spiking models under two types of noise: Gaussian distributed and rain simulation. For simplicity, only results obtained for the \texttt{Cityscapes} dataset are discussed. 

\begin{figure}[!h]
    \centering
    \includegraphics[width=0.47\columnwidth]{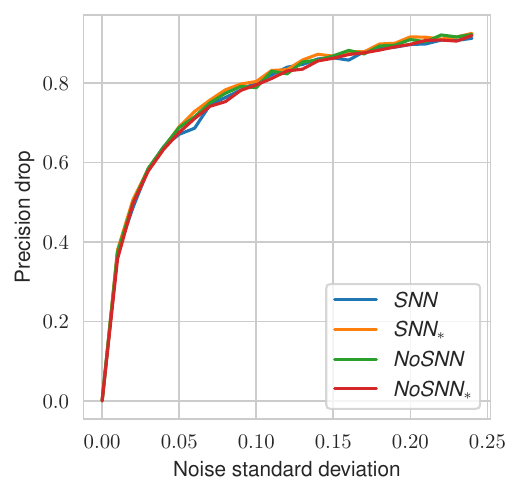}
    \includegraphics[width=0.47\columnwidth]{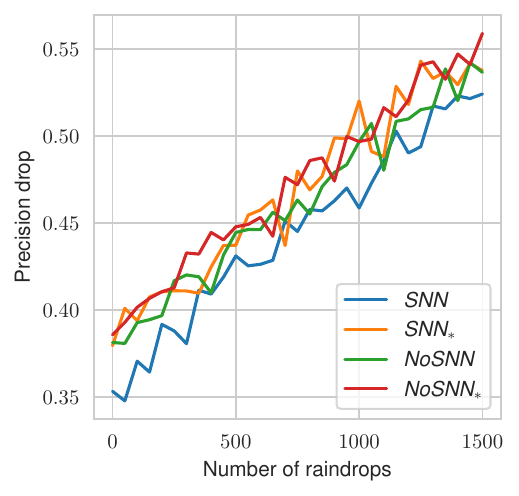}
    \vspace{-3mm}
    \caption{Relative precision decrease $(\text{mAP}_{.5} - \text{mAP}_{.5}^{noise})/\text{mAP}_{.5}$ for different intensities of Gaussian noise (left) and rain simulation (right).}
    \label{fig:noise_prec_drop}
    \vspace{-3mm}
\end{figure}

Answers to RQ3 can be found in Fig. \ref{fig:noise_prec_drop}. For the Gaussian noise, there are no significant differences between spiking and non-spiking detectors. We hypothesize that the non-spiking backbone and the direct encoding method used to convert features into spike trains does not induce as much noise as an encoder applied directly to raw image data. As for rain simulation, the non-fine-tuned SNN detector performs best by a small margin, while the rest perform similarly.

\subsection{RQ4: Can the proposed model detect unknown objects?}\label{ssec:RQ4}

The approach for detecting new objects without retraining explained in Section \ref{ssec:OWOD} hinges on a simple and intuitive heuristic: a bounding box predicted as \emph{background} that overlaps with high objectness proposals and does not overlap with predicted known objects could potentially contain a new object. Besides its low computational burden (no need for model retraining), it can be applied to any two-stage detector and requires minimum computational overhead.
\begin{figure}[!h]
\centering
\includegraphics[height=2.2cm]{000000000011.png} \includegraphics[height=2.2cm]{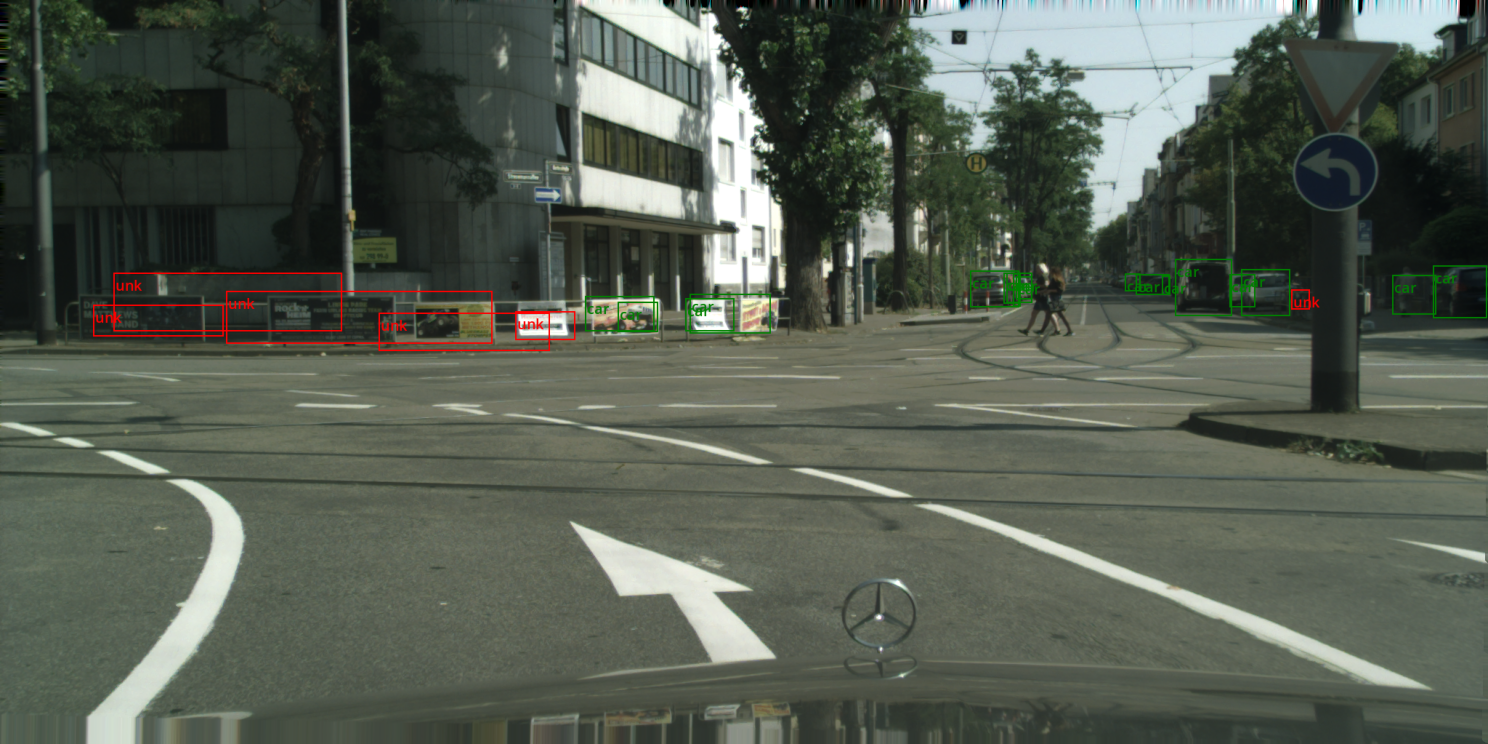}
\caption{Two example images with known (green) and unknown (red) objects detected by the proposed approach.}
\label{fig:new_objs}
\vspace{-2mm}
\end{figure}

We qualitatively analyze some examples in Fig. \ref{fig:new_objs} to point out the weaknesses of this heuristic, which relate to the reason why the detection of unknown objects is challenging when dealing with two-stage detectors: the objectness values generated by the RPN are not class-agnostic and are highly biased towards the known classes. This bias can be checked in both images. In the left figure (known classes are \emph{pedestrian}, \emph{traffic lights}, \emph{bus} and \emph{motorcycle}), there are some new objects detected near traffic signs (red bounding boxes), which actually resemble traffic signs. In the second image (known classes are \emph{car}, \emph{bus} and \emph{truck}) we observe some posters that have been detected as new objects, close to others that are mistakenly detected as \emph{cars}. The latter exposes the fact that the model is wrongly assigning high objectness to regions with posters, and therefore posters that are not predicted wrongly as \emph{cars} are detected as new objects. 

\section{Conclusions and future research}\label{sec:Conclusions}

This work has presented a directly trained fully-spiking object detector for autonomous driving. It obtains very competitive results in terms of precision and recall against non-spiking implementations of the same detector architecture, with a performance degradation that does not surpass 2\% in the worst case. By contrast, significant energy savings are reached (up to 85\% in the spiking modules). Furthermore, the performance of the proposed detector reacts similarly to non-spiking detectors when different noises are induced to the input image, showing a slight robustness increase to rain simulation when not fine-tuned. We have also designed a simple approach to detect possible new objects, which can be applied directly to two-stage detectors without retraining. The qualitative assessment of its detected objects, however, has revealed that the bias of the RPN module to known classes can assign high objectness values to objects that are similar to known classes. This observation exposes the inherent complexity of detecting new objects in two-stage detectors if retraining is to be avoided.

In the future we plan to improve the performance of the proposed detector at both training and inference phases. A major focus will be placed on extending the capabilities of these models by i) designing advanced open-set capabilities to detect new objects; and ii) characterizing and adding new classes autonomously when unknown objects appear recurrently in the scenario at hand. Fully-spiking backbones will be also investigated for the sake of a higher resiliency of the spiking detector against image artifacts.



\bibliographystyle{IEEEtran}
\bibliography{mybiblio}

\begin{thebibliography}{10}
\providecommand{\url}[1]{#1}
\csname url@rmstyle\endcsname
\providecommand{\newblock}{\relax}
\providecommand{\bibinfo}[2]{#2}
\providecommand\BIBentrySTDinterwordspacing{\spaceskip=0pt\relax}
\providecommand\BIBentryALTinterwordstretchfactor{4}
\providecommand\BIBentryALTinterwordspacing{\spaceskip=\fontdimen2\font plus
\BIBentryALTinterwordstretchfactor\fontdimen3\font minus
  \fontdimen4\font\relax}
\providecommand\BIBforeignlanguage[2]{{%
\expandafter\ifx\csname l@#1\endcsname\relax
\typeout{** WARNING: IEEEtran.bst: No hyphenation pattern has been}%
\typeout{** loaded for the language `#1'. Using the pattern for}%
\typeout{** the default language instead.}%
\else
\language=\csname l@#1\endcsname
\fi
#2}}

\bibitem{wang2021research}
P.~Wang, ``Research on comparison of {LiDAR} and camera in autonomous
  driving,'' in \emph{Journal of Physics: Conference Series}, vol. 2093, no.~1,
  2021, p. 012032.

\bibitem{muhammad2020deep}
K.~Muhammad, A.~Ullah, J.~Lloret, J.~Del~Ser, and V.~H.~C. de~Albuquerque,
  ``Deep learning for safe autonomous driving: Current challenges and future
  directions,'' \emph{IEEE trans Intell Transp Syst}, vol.~22, no.~7, pp.
  4316--4336, 2020.

\bibitem{feng2020deep}
D.~Feng \emph{et~al.}, ``Deep multi-modal object detection and semantic
  segmentation for autonomous driving: Datasets, methods, and challenges,''
  \emph{IEEE trans Intell Transp Syst}, vol.~22, no.~3, pp. 1341--1360, 2020.

\bibitem{zou2023object}
Z.~Zou, K.~Chen, Z.~Shi, Y.~Guo, and J.~Ye, ``Object detection in 20 years: A
  survey,'' \emph{Proceedings of the IEEE}, 2023.

\bibitem{singh2021order}
D.~K. Singh \emph{et~al.}, ``{ORDER}: Open world object detection on road
  scenes,'' in \emph{NeurIPS Workshops}, 2021.

\bibitem{liu2019edge}
S.~Liu, L.~Liu, J.~Tang, B.~Yu, Y.~Wang, and W.~Shi, ``Edge computing for
  autonomous driving: Opportunities and challenges,'' \emph{Proceedings of the
  IEEE}, vol. 107, no.~8, pp. 1697--1716, 2019.

\bibitem{lechner2020neural}
M.~Lechner \emph{et~al.}, ``Neural circuit policies enabling auditable
  autonomy,'' \emph{Nature Machine Intelligence}, vol.~2, no.~10, pp. 642--652,
  2020.

\bibitem{jiao2019survey}
L.~Jiao \emph{et~al.}, ``A survey of deep learning-based object detection,''
  \emph{IEEE Access}, vol.~7, pp. 128\,837--128\,868, 2019.

\bibitem{9913352}
K.~Muhammad \emph{et~al.}, ``Vision-based semantic segmentation in scene
  understanding for autonomous driving: Recent achievements, challenges, and
  outlooks,'' \emph{IEEE trans Intell Transp Syst}, vol.~23, no.~12, pp.
  22\,694--22\,715, 2022.

\bibitem{fastrcnn2015}
R.~Girshick, ``Fast {R-CNN},'' in \emph{IEEE International Conference on
  Computer Vision}, 2015, pp. 1440--1448.

\bibitem{ren2015faster}
S.~Ren, K.~He, R.~Girshick, and J.~Sun, ``Faster {R-CNN}: Towards real-time
  object detection with region proposal networks,'' \emph{NIPS}, vol.~28, 2015.

\bibitem{terven2023comprehensive}
J.~Terven and D.~Cordova-Esparza, ``A comprehensive review of {YOLO}: From
  {YOLOv1} to {YOLOv8} and beyond,'' \emph{arXiv preprint arXiv:2304.00501},
  2023.

\bibitem{transformers2020}
N.~Carion \emph{et~al.}, ``End-to-end object detection with transformers,'' in
  \emph{European Conference on Computer Vision}, 2020, pp. 213--229.

\bibitem{miller2018dropout}
D.~Miller, L.~Nicholson, F.~Dayoub, and N.~S{\"u}nderhauf, ``Dropout sampling
  for robust object detection in open-set conditions,'' in \emph{IEEE Int.
  Conf. Robot. Autom.}, 2018, pp. 3243--3249.

\bibitem{dhamija2020overlooked}
A.~Dhamija, M.~Gunther, J.~Ventura, and T.~Boult, ``The overlooked elephant of
  object detection: Open set,'' in \emph{IEEE/CVF Winter Conference on
  Applications of Computer Vision}, 2020, pp. 1021--1030.

\bibitem{openworldobjectdetection2021}
K.~Joseph, S.~Khan, F.~S. Khan, and V.~N. Balasubramanian, ``Towards open world
  object detection,'' in \emph{IEEE/CVF Conference on CVPR}, 2021, pp.
  5830--5840.

\bibitem{towardsopensetobjdetanddiscovery2022}
J.~Zheng, W.~Li, J.~Hong, L.~Petersson, and N.~Barnes, ``Towards open-set
  object detection and discovery,'' in \emph{IEEE/CVF Conference on CVPR},
  2022, pp. 3961--3970.

\bibitem{davies2018loihi}
M.~Davies \emph{et~al.}, ``Loihi: A neuromorphic manycore processor with
  on-chip learning,'' \emph{IEEE Micro}, vol.~38, no.~1, pp. 82--99, 2018.

\bibitem{akopyan2015truenorth}
F.~Akopyan \emph{et~al.}, ``Truenorth: Design and tool flow of a 65 mw 1
  million neuron programmable neurosynaptic chip,'' \emph{IEEE TCAD}, vol.~34,
  no.~10, pp. 1537--1557, 2015.

\bibitem{heeger2000poisson}
D.~Heeger \emph{et~al.}, ``Poisson model of spike generation,'' \emph{Handout,
  University of Standford}, vol.~5, no. 1-13, p.~76, 2000.

\bibitem{rathi2021diet}
N.~Rathi and K.~Roy, ``{DIET-SNN}: A low-latency spiking neural network with
  direct input encoding and leakage and threshold optimization,'' \emph{IEEE
  Trans Neural Netw Learn Syst}, 2021.

\bibitem{izhikevich2003}
E.~Izhikevich, ``Simple model of spiking neurons,'' \emph{IEEE Transactions on
  Neural Networks}, vol.~14, no.~6, pp. 1569--1572, 2003.

\bibitem{kasabov2019time}
N.~K. Kasabov, \emph{Time-space, spiking neural networks and brain-inspired
  artificial intelligence}.\hskip 1em plus 0.5em minus 0.4em\relax Springer,
  2019, ch.~4.

\bibitem{caporale2008spike}
N.~Caporale and Y.~Dan, ``Spike timing--dependent plasticity: a {Hebbian}
  learning rule,'' \emph{Annu. Rev. Neurosci.}, vol.~31, pp. 25--46, 2008.

\bibitem{zenke2018superspike}
F.~Zenke and S.~Ganguli, ``Superspike: Supervised learning in multilayer
  spiking neural networks,'' \emph{Neural Computation}, vol.~30, no.~6, pp.
  1514--1541, 2018.

\bibitem{cordone2022object}
L.~Cordone, B.~Miramond, and P.~Thierion, ``Object detection with spiking
  neural networks on automotive event data,'' in \emph{International Joint
  Conference on Neural Networks}, 2022, pp. 1--8.

\bibitem{kim2020spikingyolo}
S.~Kim, S.~Park, B.~Na, and S.~Yoon, ``Spiking-yolo: spiking neural network for
  energy-efficient object detection,'' in \emph{AAAI Conference on Artificial
  Intelligence}, vol.~34, no.~07, 2020, pp. 11\,270--11\,277.

\bibitem{kim2022beyond}
Y.~Kim, J.~Chough, and P.~Panda, ``Beyond classification: Directly training
  spiking neural networks for semantic segmentation,'' \emph{Neuromorphic
  Computing and Engineering}, vol.~2, no.~4, p. 044015, 2022.

\bibitem{barchid2022spiking}
S.~Barchid, J.~Mennesson, J.~Eshraghian, C.~Dj{\'e}raba, and M.~Bennamoun,
  ``Spiking neural networks for frame-based and event-based single object
  localization,'' \emph{arXiv preprint arXiv:2206.06506}, 2022.

\bibitem{sengupta2019going}
A.~Sengupta, Y.~Ye, R.~Wang, C.~Liu, and K.~Roy, ``Going deeper in spiking
  neural networks: {VGG} and residual architectures,'' \emph{Frontiers in
  neuroscience}, vol.~13, p.~95, 2019.

\bibitem{maskrcnn}
K.~He, G.~Gkioxari, P.~Dollár, and R.~Girshick, ``Mask {R-CNN},'' in
  \emph{International Conference on Computer Vision}, 2017, pp. 2980--2988.

\bibitem{cordts2016cityscapes}
M.~Cordts \emph{et~al.}, ``The cityscapes dataset for semantic urban scene
  understanding,'' in \emph{IEEE Conference on CVPR}, 2016, pp. 3213--3223.

\bibitem{varma2019idd}
G.~Varma, A.~Subramanian, A.~Namboodiri, M.~Chandraker, and C.~Jawahar,
  ``{IDD}: A dataset for exploring problems of autonomous navigation in
  unconstrained environments,'' in \emph{IEEE Winter Conference on Applications
  of Computer Vision}, 2019, pp. 1743--1751.

\bibitem{yu2020bdd100k}
F.~Yu \emph{et~al.}, ``Bdd100k: A diverse driving dataset for heterogeneous
  multitask learning,'' in \emph{IEEE/CVF Conference on CVPR}, 2020, pp.
  2636--2645.

\bibitem{pehle2021norse}
C.~Pehle and J.~E. Pedersen, ``Norse--a deep learning library for spiking
  neural networks,'' \emph{https://doi.org/10.5281/zenodo.4422025}, 2021.

\bibitem{lin2014coco}
T.-Y. Lin \emph{et~al.}, ``Microsoft {CoCo}: Common objects in context,'' in
  \emph{European Conference on Computer Vision}, 2014, pp. 740--755.

\bibitem{computing2014energy}
M.~Horowitz, ``1.1 computing's energy problem (and what we can do about it),''
  in \emph{IEEE International Solid-State Circuits Conference Digest of
  Technical Papers}, 2014, pp. 10--14.

\end{thebibliography}

\end{document}